# A Perspective on Robotic Telepresence and Teleoperation using Cognition: Are we there yet?

Hrishav Bakul Barua[1], Ashis Sau[1], Ruddra dev Roychoudhury[1]

*Abstract*— Telepresence and teleoperation robotics have attracted a great amount of attention in the last 10 years. With the Artificial Intelligence (AI) revolution already being started, we can see a wide range of robotic applications being realized. Intelligent robotic systems are being deployed both in industrial and domestic environments. Telepresence is the idea of being present in a remote location virtually or via robotic avatars. Similarly, the idea of operating a robot from a remote location for various tasks is called teleoperation. These technologies find significant application in health care, education, surveillance, disaster recovery, and corporate/government sectors. But question still remains about their maturity, security and safety levels. We also need to think about enhancing the user experience and trust in such technologies going into the next generation of computing.

## I. INTRODUCTION

People have envisioned remote presence and operation since long past. The idea of being present in a remote location virtually and performing certain actions is no longer a fiction. The effects of Telepresence [1], [2] and Teleoperation can be seen greatly in the recent times. The most promising scenario considers representing a person in a remote location for meetings, seminars, conferences, health consultancies (Covid-19 scenarios [3]), teaching etc. being his avatar [4]. But the main concern to fully realize such an avatar robot depends upon its underlying architecture which needs to be robust, efficient and real-time at the same time [5].

Both Telepresence and Teleoperation robots allow a human to be present and act in geographically separated locations via communication networks. The human can also perceive and interact with the physical entities of the remote environment. This actually requires a high fidelity and accurate geometry based 3D representation of the remote environment in the user's end. Apart from this the robot also requires a robust navigation approach for autonomous maneuvering with minimum user inputs. Another important aspect is the Human-robot interaction capabilities such as speech recognition, gesture recognition and visual cues understanding. How well the robot adheres to societal norms and regulations while interacting with people and navigating through social spaces is an important question. Although much research has been conducted in all of the mentioned areas but question still remains about the adequacy. In this short article we put forward some of the arguments, challenges and prospects for both Telepresence and Teleoperation robotics from a practical viewpoint. Figure 1 shows a trend of search interest for terms (related to cognitive robotics) over time retrieved from Google Trends application. The data is from 2004 January till 2021 July. The X-axis signifies the absolute search volume and Y-axis is time. It is seen that AI and visual perception are constantly gaining popularity compared to NLP and augmented reality. More focus is being given in vision side of cognitive robotics whereas NLP is somewhat neglected as per the search report. Augmented reality is not less searched but not too highly as well. So, researchers are achieving more success in vision related applications to robotics than other domains. More focus is needed in improving NLP and augmented or virtual reality applications to achieve a full fledged cognitive sense in telepresence and teleoperation robotics.

[1] H. B. Barua, A. Sau, and R. D. Roychoudhury are with the Robotics & Autonomous Systems Research group (Cognitive Robotics) of TCS Research, Kolkata, India.

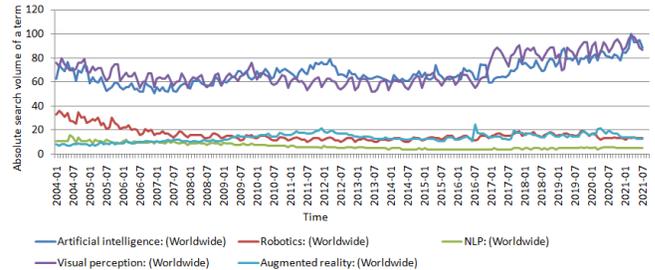

Fig. 1: Popularity trends of terms related to cognitive robotics and as retrieved for Google Trends.

## II. CLOUD-CENTRIC, EDGE-CENTRIC OR HYBRID MODELS FOR COMPUTATION?

The amount of computation required for a Telepresence or Teleoperated robot is significant. The tasks like Simultaneous Localization and Mapping (SLAM) for exploration and navigation, Computer vision tasks like object recognition, 3D scene reconstruction and understanding, communication related computations, human-robot interaction related processing etc. demand a huge amount of compute and storage resources. The Cloud-centric models handle the compute and storage requirements quite well as almost all of the tasks are offloaded to the cloud, but this leads to excessive network overheads negatively impacting the real-time execution. Similarly, edge-centric models handles the real-time requirement by minimizing the network overheads but compromises on the compute/storage capabilities impacting the output quality. So, it is necessary to use the best of both these worlds in designing robots, such that the most heavy computations (which also have a lower real-time response requirement) can be offloaded to the cloud and the other normal computations may be accomplished in the edge itself. A hybrid cloud-edge centric system is the solution we see at this point.

The main challenge in a hybrid cloud-edge centric approach is how the system distributes its tasks between cloud and edge. One easy holistic approach can be the distribution of high resource demanding tasks like audio-video communication into cloud-based approach whereas edge-based approach takes the control of the tasks where reliability and realtimeness are the key factors like robot control systems or haptic response based tasks. One similar architecture and solution has been discussed in [1]. Figure 2 depicts a holistic architecture of such kind of telepresence and teleoperation system with cognitive capabilities.

## III. REAL-TIME IS THE KEY

Real-time requirements form a major part of the success story when it comes to robots representing a human in a remote location. A constant communication between the human and the robot is indispensable. The time critical applications like audio-visual transfer between the user and the robot, navigational instructions and interaction commands (such as speech and visual cues), require a robust low latency based communication protocol (over UDP) on the application layer itself. In case of visual data being sent from user's end to robot's end, sometimes the video may not be transferred correctly and fully if the bandwidth and network strength is low. For example, if the face of the user is to be displayed in the robot end, so instead of transferring the video directly, the face can be reconstructed in the robot end using 3D rendering techniques which would require transfer of some meta data instead. Such a protocol is being described in [1], where a unique acknowledgement scheme is introduced. Acknowledgements are required for messages with high reliability requirement. For rest of the messages, a best-effort way is adopted to deliver messages as mentioned in [6]. Similar approach has also been extended along with a novel packetization and rendering technique in [6] for video transmission which improves the perceived quality of service (QoS) of the visual-feed.

## IV. SECURITY, PRIVACY AND RELIABILITY

Since we are speaking about data transfer between user and robot and vice versa, which includes cloud and edge platforms via internet technologies and protocols, it is important to look at the security, privacy and reliability aspects. Since, cloud is an important player in a hybrid model, security as a service [7] and privacy as a service [8] can be adopted for keeping the private data secure at an organizational level. Reliability is another important player in this arena. Transfer of data from the user to the robot is important but more crucial is how reliably it is been done. Here we can have a context aware reliability model to simplify the requirements in accordance with real-time capabilities. The data packets which are important with respect to the overall success of the telepresence/teleoperation tasks need to be sent in high reliability mode and rest can be sent in low reliability setup. For example, a command by the user to perform an important task can be sent in high reliability mode but a simple move forward command can be ignored, which can be later compensated with autonomous navigation capabilities.

| Related works | Social Awareness and Intelligence | Real-time planning in Navigation of Goals | Distributed Computation | Multi-modal Interfacing |
|---|---|---|---|---|
| Shiarlis et al. Paper [9] | ✓ | ✓ | ✗ | ✗ |
| Cosgun et al. Paper [10] | ✓ | ✓ | ✗ | ✗ |
| Budiharto et al. Paper [11] | ✗ | ✗ | ✗ | ✗ |
| Wang et al. Patent [5] | ✗ | ✗ | ✗ | ✗ |
| Taranovic et al. Paper [12] | ✗ | ✗ | ✗ | ✗ |
| Paper [1] | ✓ | ✓ | ✓✓ | ✓✓ |

TABLE I: Comparison of related works on the basis of high level features.

## V. SOCIAL AWARENESS BASED INTELLIGENCE

Next thing to look at is the behaviour of the robot in a social setup. The acceptance and success of a telepresence and teleoperation robot largely depends on its social behaviour and intelligence. The major scenario being how well the robot navigates in a social environment between humans and other moving objects in a natural manner. This requires the robot to predict the future position of the moving body (human or non-human) with certain accuracy [13]. Another scenario is the capability of a robot to go and join groups of people [14], [15], [16] in random scenarios for meetings and discussions and also paying attention to a speaking person in the group [2], [17]. There are many such scenarios which need to be considered for a robot to make it befitting for a social environment and more human like.

## VI. COGNITIVE CAPABILITIES

A human can simply make his/her presence in a geographically different location through a audio-video call/conference. But it does not reflect all human capabilities there. The robotic era gives us the opportunity to overcome this. Now the robotic world currently moving from a simply telepresence towards a teleoperation era by its cognitive capabilities.

A human can perceive the environment with eyes, ears, nose, tongue and skin. Finally, the human processes all the perceived information in the brain and take necessary actions which is called cognition. Similarly, a robot can perceive the environment through vision (visual perception), speech (NLP) and touch inputs using various sensors. Much success has been accomplished in the field of computer/robot vision like object detection, obstacle recognition, human recognition and layout prediction but the amount of accuracy and fidelity required is still not up to the mark. Speech and Natural Language Processing (NLP) systems are far lagging behind than required for a robot to interact in a non constraint environment. Researchers are going on to make robots understand complex human instructions and plan their actions accordingly [23], [24]. Other works focus on dialogue based systems for human-robot interaction and vision guided instruction understanding from scenes and speech. Real-time planning of paths and navigational goals also require on the fly cognitive capabilities. The perception of touch using pressure and tactile sensors have come to prominence in the last few years. It is an important aspect in tele robotics as to understand the environment by the remote user and perform actions via the robot using haptic gloves. For example while holding a heavy material, the robot needs to exert more pressure but for a glass it needs to regulate its pressure else the glass could break.

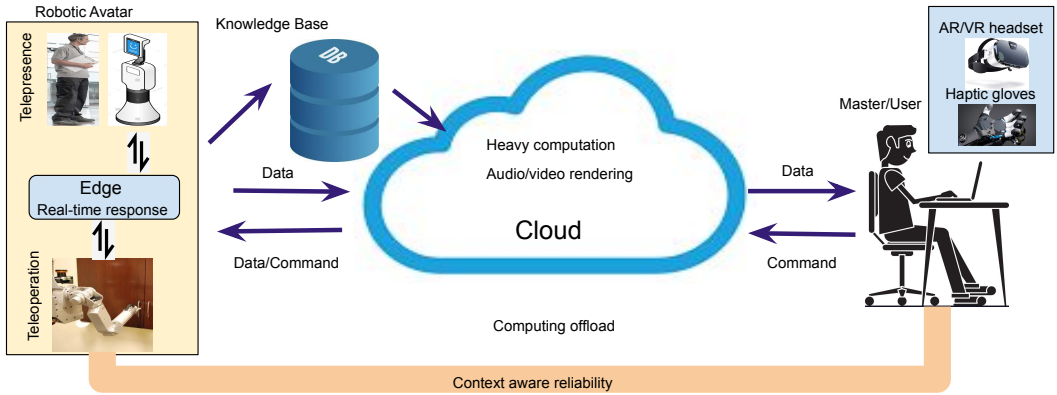

Fig. 2: Architecture of a distributed edge-centric telepresence system using a mobile robot.

| Related works | Cloud Centric | Edge Centric | Computational bandwith distribution between edge and cloud | Degree of realtimeness | Privacy of Information | Context aware reliability |
|---|---|---|---|---|---|---|
| Chen et al. Paper [18] | ✓ | ✓ | ✗ | High | High | ✗ |
| King et al. Paper [19] | ✓ | ✗ | ✗ | High | Medium | ✗ |
| Bickmore et al. Patent [20] | ✓ | ✗ | ✗ | Medium | Medium | ✗ |
| Zhani et al. Paper [21] | ✓ | ✓ | ✓ | High | High | ✗ |
| Ghosh et al. Paper [22] | Implementable in Cloud | Implementable in Edge | NA | High | High | ✗ |
| Paper [1] | ✓✓ | ✓✓ | ✓ | High | High | ✓ |

TABLE II: Comparison of related works on the basis of technical and communication related features.

## VII. Multiple mode of operation: Automatic and Manual

Tele robotics can operate in two modes basically- Manual and Automatic. In manual mode, the robot performs exactly as instructed by the user in the remote location, be it navigation, path planning, picking and placing, following someone or so on. In automatic mode the robot performs almost all the activities with its own intelligence and capability. Both have their merits and demerits. In manual mode, the user is in full control of the robot and its activities but the main challenge is the environment. Sometimes the user's perception of the remote environment may be erroneous or incorrect, which can lead to unwanted robotic activity. For example, the distance from the robot to a person in scene may be wrongly judged by the user eventually making the robot bump into the user. In autonomous mode, the full control is with the robot and the robot perceives the environment and acts as per its intelligence. But the issue is if a module fails in the robot, it might lead to unwanted behaviour of the robot, even hurting people in the scene. Moreover, in manual mode, if there is an issue with the communication network the user command may not be transmitted to he robot reliably. So, recent researches are focused on a shared autonomy kind of a logic where the robot is not fully automatic and the user can choose among manual and automatic modes as per the need of the situation. Such a hybrid modality can enhance user's experience as well as robots perception from the viewers point. The co-located people can have a better experience interacting with the tele robot and feel safe around it.

## VIII. The future is Virtual/Augmented Reality?

In Tele robotics, the user gets the view of the remote location in the control screen usually in direct streaming video or 3D reconstructed scene format. But, is it enough? We already mention one of the issues, where perceiving the environment incorrectly by the remote user can lead to unwanted robotic activity. This perception can be improved using immersive technology like Virtual reality (VR) and Augmented reality (AR). The complete real-life 3D reconstruction of the remote location with all the living and non-living objects accurately represented in the user's end can mitigate this challenge. The user can have a headset and a set of haptic gloves to experience and interact with the virtual/augmented environment and its objects with high accuracy.

Few research works are currently focused on the enhancement of robotic cognitive capabilities by augmentation in the remote constraint environment. In an organizational teleconference scenario, the remote person can control his/her robotic avatar by navigating based on the visual feedback from it. Also, he/she can control/maneuver and track the avatar using the floor-map of the remote environment when available.

## IX. IMPACT IN DIFFERENT USE CASES

Robotic telepresence/teleoperation can produce enormous impact towards society modernization. A generic telepresence/teleoperation robotic solution can be very efficient in different application scenarios. Here, we put some light on its applicability in some selected use-cases.

Robotic avatar in agile remote meetings/conferences: Now-a-days we attend meetings from different geographic locations using audio-video conferencing system. But a robotic solution makes it more agile and interactive considering the cognitive capabilities. A robotic solution can remotely connect people more efficiently saving time. This also reduces cost incurred by conveyance, food, and stay while conferences are organised abroad.

Remote tele-inspection in hospitals: The telepresence/teleoperation solution for remote inspection in hospitals is the next big thing. A robot may not be involved in critical operations in recent times,but it can provide live guidance from a remote doctor when the robotic avatar is present in the operation theatre. Also it can be used for remote inspection by a doctor (who is not physically present at the remote location) in different health-care centers at rural areas. A similar integrated architecture and prototype has been discussed in [1] for isolation wards in hospitals. This solution has high impact during current COVID-19 pandemic situation [3]. We can ensure safety of a health personnel from infections by mitigating the infection spreading rate.

Tele-education in schools/colleges: A robotic solution will open a new era in tele-education by representing a specialized teacher/professor remotely in schools/colleges at under-privileged areas where he/she may not be able to be present physically. Remote tele-surveillance: A telepresence architecture can be implemented over drone as well to be used by police for remotely monitoring and inspecting in crime prone areas.

Remote guide in libraries: In libraries, a robotic avatar can navigate across the book racks and take note of the existing titles on behalf of a person to cut down his/her efforts and time.

A telepresence robot can play a huge role for emotional support in different elderly-care, and children's home. Figure 3 shows the popularity trends of tele robotics w.r.t. application areas and use cases.

## X. DISCUSSIONS

We put forward some of the recent and relevant concern areas in the field of telepresence and teleportation technology with a certain perspective in mind. We consider both high level and low level features in our discussion. We envision the drawbacks and future capabilities that require attention and development both in research as well as implementation domain. Table I gives a concise comparison of current works (in the form of papers and patents) on the basis of high level features as discussed in the previous sections. Similarly Table II pinpoints the comparative study between recent papers/patents on the basis of low level technical features (also concisely discussed in the previous sections). In both the tables, ✗- signifies no presence of that feature in that work, ✓- signifies presence but not fully matured and ✓✓- means full maturity of that feature. Analyzing the data from the tables it is evident that although the features discussed in the paper are present in some of the current works but are not fully matured. Specially, intelligence and cognition of a robot pertaining to social awareness and human-robot interaction is missing. Real-time reaction to environmental change and adaptability is low in current works. Harnessing the full power of cloud-edge systems for robotics by computational load distribution is not satisfactory. Reliability is present in the current systems but not in its full potential. Finally, we put forward a list of most popular telepresence robots in the market today in Table III.

## XI. INITIATIVES

The XPRIZE foundation [1], which encourages people to develop technology for humanity by hosting competitions, has announced $10M ANA Avatar XPRIZE [2]. Their main aim is to encourage the design and development of a tele robot that can see, hear, touch and interect like a human and bridge the geographic gap between two distant locations. A human can be virtually transferred to a distant location in real-time where his expertise and skills are needed. It was launched in 2018. Recently, the 38 semi-finalist teams from 16 different countries are announced[3]. After a series of competitions, the winners will be declared in 2022. The team[4] from TCS Research (in collaboration with IISc Bangalore, Hanson Robotics and ARTPARK) is the only team from India to be in the semi-finals. A demonstration of this telepresence and teleoperated humanoid robotic nurse can be seen in the project website[5]. Figure 4 shows the tele robot at its full scale. Figure 5 shows the robot performing grasping operation (holding and taking a water glass from a table) and delivering operation in interaction with a human. The tele robot "Asha" is capable of communicating with humans in a socially interactive manner [25], [26]. Asha uses both audio-visual as well as gesture cues for conveying information as well as perceiving it. We need more such initiatives to encourage the design and development of cutting edge tele robots for our future generation. Apart from these challenges, we have top notch conferences like IEEE International Conference on Robotics and Automation (ICRA), IEEE/RSJ International Conference on Intelligent Robots and Systems (IROS) and ACM/IEEE International Conference on Human-robot Interaction (HRI) which can ensure a bright and fulfilling future for Telepresence/Teleoperation robotics.

---

[1] https://en.wikipedia.org/wiki/X_Prize_Foundation
[2] https://www.xprize.org/
[3] https://www.xprize.org/prizes/avatar/competing-teams
[4] https://aham-avatar.org/
[5] https://aham-avatar.org/demo/

| Telepresence Robot | Application areas | Technical features | Available sensors | References |
|---|---|---|---|---|
| Double 3 by Double Robotics 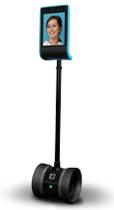 | Generic remote conference | <ul><li>WebRTC based (128-bit AES encrypted) video conference</li><li>Click-to-Drive Mixed Reality UI</li><li>Motorized height control</li><li>Lateral Stability Control (LSC)</li></ul> | <ul><li>2 x 13 MP Unified Pan/Tilt/Zoom Cameras (one super wide angle lens, one super zoom lens) with Night Vision Mode</li><li>2 x Stereovision depth sensors (Intel® RealSense™ D430)</li><li>5 x Ultrasonic range finders</li><li>2 x Wheel encoders (2048 PPR each)</li><li>1 x Inertial Measurement Unit (9 DoF)</li></ul> | https://www.doublerobotics.com/ |
| Ohmni Robot by OhmniLabs 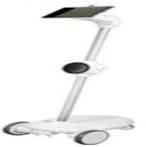 | Tele-education, Health-care, Senior-care | <ul><li>Advanced vision based Autodocking Technology</li><li>Glide Drive Technology for smooth motion on any surface</li><li>Low-latency video encoding system</li></ul> | <ul><li>13MP Wide Angle Camera</li><li>LiDAR, depth cameras</li></ul> | https://ohmnilabs.com/ |
| Ava by Ava Robotics 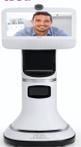 | Corporate, Health-care, Hospitality | <ul><li>Cisco Webex based video conferencing system</li><li>Autonomous navigation based on specifying destination</li><li>Built-in collision avoidance technology</li><li>Automatic self-docking</li></ul> | <ul><li>Three Cameras including a 360-degree Camera</li><li>3D depth camera, LiDAR, inertial measurement unit</li></ul> | https://www.avarobotics.com/ |
| Beam by Suitable Technologies 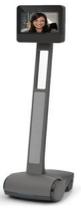 | Health-care, Tele-education, Corporate, Manufacturing, Cultural Arts, Retail | <ul><li>Increase sound clarity by duplexing mic-array inputs and eliminating background noise and echoes</li><li>Efficient communication using Beam's peer-to-peer UDP connections along with AES-256 based encryption, authentication(HMAC-SHA1) and random number-generated keys to optimize privacy and security</li></ul> | <ul><li>Array of three forward and one rear facing microphones</li><li>Two HDR cameras</li></ul> | https://suitabletech.com/products/beam/ |
| VGo by Vecna 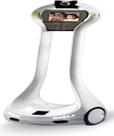 | Health-care | <ul><li>Upto 5x video zoom feature along with hi-resolution photos for video conferencing</li><li>Mouse and keyboard control for navigation and camera tilting</li></ul> | <ul><li>Cliff sensor</li></ul> | https://vecnahealthcare.com/vgo/ |
| Anette Telepresence Robot 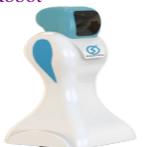 | Generic remote conference | <ul><li>Autonomous navigation, obstacle avoidance, cliff avoidance, autodocking facilities with real-time video conferencing</li></ul> | NA | https://sastrarobotics.com/products/anette-telepresence-robot/ |
| BotEyes Telepresence Robot 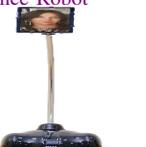 | Remote conferences, Inspection in manufacturing units, Tele-education | <ul><li>Video conferencing tool: WebRTC (default), Skype or others</li><li>Autodocking and Obstacle avoidance facility</li><li>160 deg robot-head tilt angle to look the floor and ceiling</li></ul> | <ul><li>infrared sensors, ultrasonic distance sensors for obstacle avoidance, autodocking</li><li>orientation sensor in tablet for autodocking</li></ul> | https://boteyes.com/ |

TABLE III: List of various telepresence robots available in the market today with their applicability, features, sensors and references.

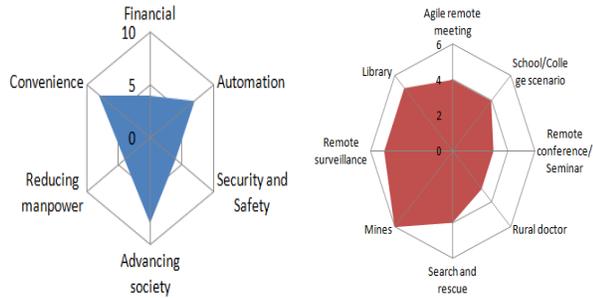

Fig. 3: The left side image shows application wise popularity radar graph of telepresence and teleoperation technologies. The right hand side image shows a similar graph for different use cases.

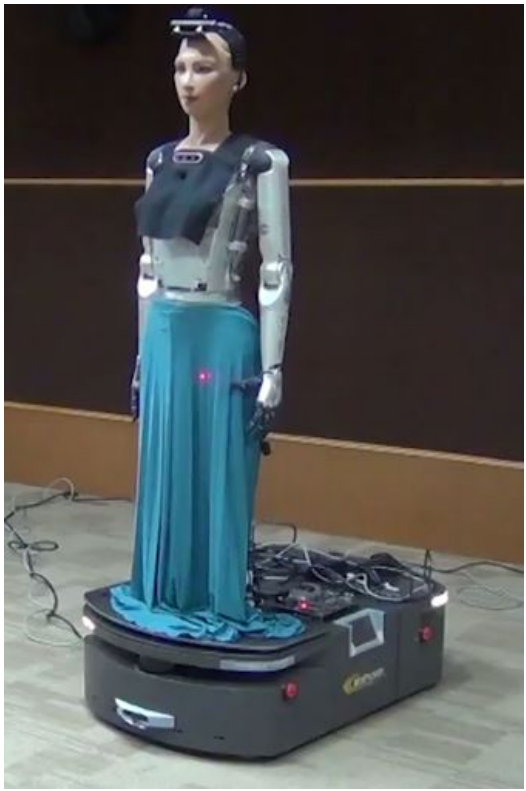

Fig. 4: Full scale image of the robotic avatar "**Asha**" for Telepresence and Teleoperation.

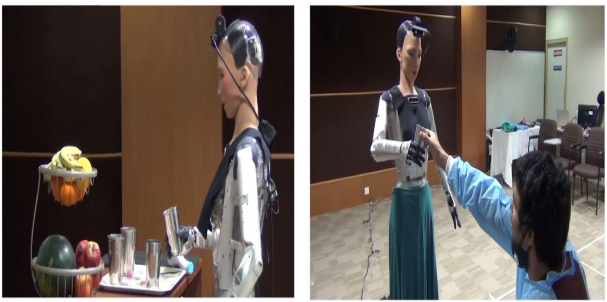

Fig. 5: Images of the Telepresence and Teleoperated robot "**Asha**". Left side: Grasping water glass from table. Right side: Handing the water glass to a human.